\documentclass[runningheads]{llncs}
\usepackage[T1]{fontenc}
\usepackage{graphicx}
\graphicspath{{img/overview/}}
\usepackage{calc}
\usepackage{subfig}
\usepackage{amsfonts}
\usepackage{hyperref}
\usepackage{color}

\usepackage{colortbl}
\usepackage{dsfont}
\usepackage{tabularx}
\usepackage{booktabs}
\usepackage{mathtools}
\usepackage{todonotes}
\usepackage{xargs} 
\usepackage{multirow}
\usepackage{multicol}
\usepackage[misc]{ifsym}

\usepackage[]{todonotes}
\newcommandx{\unsure}[2][1=]{\todo[linecolor=red,backgroundcolor=red!25,bordercolor=red,#1]{#2}}

\newcommandx{\tbd}[2][1=]{\todo[linecolor=red,backgroundcolor=red!25,bordercolor=red,inline,#1]{TBD: #2}}

\newcommandx{\resp}[2][1=]{\todo[linecolor=red,backgroundcolor=red!25,bordercolor=red,#1]{Responsible: #2}}

\newcommandx{\sebastian}[2][1=]{\todo[linecolor=green,backgroundcolor=green!25,bordercolor=green,inline,#1]{Sebastian: #2}}

\newcommandx{\juliane}[2][1=]{\todo[linecolor=green,backgroundcolor=green!25,bordercolor=green,#1]{Juliane: #2}}

\newcommandx{\mf}[2][1=]{\todo[linecolor=red,backgroundcolor=green!25,bordercolor=red,inline,#1]{MF: #2}}

\newcommandx{\karim}[2][1=]{\todo[linecolor=green,backgroundcolor=green!25,bordercolor=green,#1]{Karim: #2}}

\begin{document}
\title{Counterfactual-based Root Cause Analysis for Dynamical Systems}
%
%
\author{Juliane Weilbach \Letter \inst{1,2} \and
Sebastian Gerwinn\inst{1} \and
Karim Barsim \inst{1}\and 
Martin Fränzle \inst{2}}
\tocauthor{Juliane Weilbach, Sebastian Gerwinn, Karim Barsim, Martin Fränzle}
\toctitle{Counterfactual-based Root Cause Analysis for Dynamical Systems}
%
\institute{Bosch Center for Artificial Intelligence \and
Carl von Ossietzky University of Oldenburg}

\maketitle              
\begin{abstract}
Identifying the underlying reason for a failing dynamic process or otherwise anomalous observation is a fundamental challenge, yet has numerous industrial applications. Identifying the failure-causing sub-system using causal inference, one can ask the question: "Would the observed failure also occur, if we had replaced the behaviour of a sub-system at a certain point in time with its {\em{normal}} behaviour?" To this end, a formal description of behaviour of the full system is needed in which such counterfactual questions can be answered. However, existing causal methods for root cause identification are typically limited to static settings and focusing on additive external influences causing failures rather than structural influences. In this paper, we address these problems by modelling the dynamic causal system using a Residual Neural Network and deriving corresponding counterfactual distributions over trajectories. We show quantitatively that more root causes are identified when an intervention is performed on the structural equation and the external influence, compared to an intervention on the external influence only. By employing an efficient approximation to a corresponding Shapley value, we also obtain a ranking between the different subsystems at different points in time being responsible for an observed failure, which is applicable in settings with large number of variables. We illustrate the effectiveness of the proposed method on a  benchmark dynamic system as well as on a real world river dataset.
\keywords{Dynamic Root Cause Analysis  \and Counterfactual Inference \and Dynamic Systems.}
\end{abstract}

\section{Introduction}
Explaining unexpected behaviour in terms of underlying causes is a difficult challenge with a broad range of applications. Such applications range from  identifying potential problems in industrial processes to understanding influencing factors in anomalous weather phenomena. For example, within an assembly line of an industrial manufacturing plant, faster identification of root causes of increased scrap rate (the rate at which assembled products fail quality assessment audits) can minimize cost, increase production yield, and increase overall efficiency. 
If one can observe sufficiently many instances of anomalous behaviour or of faulty traces of a process, one option would be to perform correlation based analysis or causal discovery \cite{Spirtes2000}, thereby estimating the influencing factors to the variable "fault" \cite{wwwaw,NEURIPS2022_c9fcd02e,pmlr-v206-assaad23a}.
Alternatively, causal inference can be used even if only a single anomalous observation is available \cite{strobl2022identifying,pmlr-v162-budhathoki22a}. Here, the identification of root causes is formulated in terms of a counterfactual query:
"Would the observed failure also occur, if we had replaced the behaviour of a sub-system at a certain point in time with its {\em{normal}} behaviour?". Although such a causal inference approach can estimate a ranked score of each variable involved of being the underlying root cause, we address three main shortcomings of this approach in this paper:\\
\textbf{Static systems:} Root cause analysis based on causal inference has been considered only in static environments \cite{pmlr-v130-budhathoki21a,strobl2022identifying,pmlr-v162-budhathoki22a}. To address this limitation, we fit  a time-discretized version of an Ordinary Differential Equation (ODE) system, thereby obtaining  a dynamic model. By deriving counterfactual distributions over trajectories we then employ similar strategies as in the static case.\\
    \textbf{Structural influences:} Existing causal inference methods using counterfactuals \cite{strobl2022identifying,pmlr-v162-budhathoki22a} focus on additive external influences causing failures rather than structural influences. While \cite{pmlr-v206-assaad23a} also considers structural influences, the method is limited to linear models and does not include single time external influences. In this paper, we address this problem by allowing for interventions on the structural equation and the external influence.\\
    \textbf{Non-linear systems:} Existing  methods for root cause analysis are typically limited to linear dynamic models. Here, we address this problem by allowing transition functions to be non-linear using a simple neural network architecture.
Additionally, existing methods are limited to small systems as they rely on the computation of Shapley values, which scales exponentially with the number of variables. This becomes infeasible in a dynamic setting, since the corresponding causal graph -- unrolled over time -- would have an increasingly large number of nodes. While approximate methods for the computation of Shapley values have been proposed \cite{lundberg_unified_2017},  we suggest a simple approximation to the Shapley value, which is applicable in settings with large number of variables. \\
The remainder of this paper is organized as follows: in Section \ref{sec:related_work}, we review the related work mentioned above in more detail, and provide necessary background and notation in Section \ref{sec:background}. In Section \ref{sec:identification}, we describe our method for identifying root causes. In Section \ref{sec:experiments}, we first illustrate the mechanisms of the proposed method in a synthetic linear and non-linear setting before evaluating it on a benchmark dataset of \cite{pmlr-v206-assaad23a} as well as on real data describing river levels as in \cite{pmlr-v162-budhathoki22a}. In Section \ref{sec:conclusion}, we conclude the paper.





\section{Related Work}
\label{sec:related_work}
The problem of identifying the root cause of a system failure or anomaly has been addressed in various domains, including healthcare \cite{strobl2022identifying}, financial income distributions \cite{RCA-unit-level}, reliability engineering \cite{NEURIPS2022_c9fcd02e}, to name a few.
In the context of time-series data, causal inference techniques have been used to qualitatively explain outliers using counterfactual trajectories \cite{sulem2022diverse}. To detect root causes affecting graphical structure or transition function Assad et. al. \cite{pmlr-v206-assaad23a} propose a method based on assessing the direct causal effect. Modelling such causal effect with linear models, Assad et. al. \cite{pmlr-v206-assaad23a} show that the total effects change if the underlying causal model changes. In turn, they can use this fact to identify structural changes in the causal model. However, the method is limited to linear models and does not include single time external influences. If more observations of anomalous data are available, the problem of identifying the root causes is also amenable to statistically estimate the correlation or causation  of the different variables and time points onto the variable associated with the label "anomalous".
To this end, Tonekaboni et. al. \cite{wwwaw} introduce \underline{f}eature \underline{i}mportance in \underline{t}ime (FIT), a scoring mechanism to quantify importance of features in a multi-variate time-series.
The authors propose to assess feature importance based on their predictive power w.r.t.\ the outcome distribution, while accounting for temporal distributional shifts.
The approach localizes important features over time and can thus be used to gain useful insights into the behaviour of dynamic systems.
However, FIT does not leverage the causal structure of the underlying system and rather provides correlative explanations for the observed outcomes. \\
Best aligned with our approach, though, is the work by \cite{pmlr-v162-budhathoki22a} which defines the problem of identifying root causes of a system failure as a counterfactual query.
With this reformulation, the authors claim to be the first to propose actionable explanations to anomalous behaviour of underlying systems.
In principle, counterfactual reasoning assumes, and leverages, complete causal knowledge of the underlying system in the form of a \underline{s}tructural \underline{c}ausal \underline{m}odel (SCM).
More precisely, the work in \cite{pmlr-v162-budhathoki22a} assumes \emph{invertible} functional causal models: models in which exogenous variables are computable from endogenous system observations.
In fact, the authors leverage the default split between endogenous and exogenous variables in a graphical causal model to disentangle a node's inherited impact from its own contribution.
They account for the notion of graded causation \cite{halpern2015graded} and provide order-independent feature scoring using a game-theoretic concept commonly adopted in explainable machine learning \cite{lundberg_unified_2017}, namely Shapley values \cite{shapley1953value}. 
With its computation complexity, their approach lacks direct applicability to dynamical systems. In our experiments, we compare against the linear model performing interventions on the exogenous variable analogously to \cite{pmlr-v162-budhathoki22a}.

\section{Background and Notation}
\label{sec:background}
As mentioned in the introduction, we are interested in a counterfactual approach to identify the root cause of a system failure. In this section, we introduce the necessary concept from the literature and also introduce the notation we use throughout the paper.
Following the notation from Peters et al. \cite{PetJanSch17}, we denote the sequence of observations of the system of interest by \textit{d}-variate time series $(\textbf{Y}_t)_{t \in \mathbb{Z}}$ where each $\textbf{Y}_t$ for fixed $t$ is the vector $(Y^1_t, ...,Y^d_t)$. Each $Y_t^j$ represents the \textit{j}th observable of a system at time $t$.
By some abuse of notation, if we omit super- or subscripts, we refer to the full time series. That is, $\textbf{Y} = (\textbf{Y}_t)_{t \in \mathbb{Z}}$, $\textbf{Y}^j = (Y^j_t)_{t \in \mathbb{Z}}$ and $\textbf{Y}_t = (Y^j_t)_{j \in \{1,\dots,d\}}$.
The full time causal graph $\mathcal{G}_t$ with a node for each time point and signal $Y^j_t$ for $(j,t) \in {1,...,d} \times \mathbb{Z}$ has theoretically infinitely many nodes and is assumed to be acyclic, while the summary graph $\mathcal{G}$ with nodes $Y^1,...,Y^d$ may be cyclic.  

\begin{definition}
(Structural causal model (SCM)) \cite{PetJanSch17}\\
An SCM $\mathcal{M}({\mathcal{S}}, P_N, \mathcal{G})$ is defined by a set of structural equations $\mathcal{S}$, an acyclic graph $\mathcal{G}=(\mathcal{Y},\mathcal{E})$, and a set of independent noise variables $N^j\sim P_{{N}^j}, j\in \mathcal{Y}$. The structural equations for each node $j$ are given by:
\begin{align*}
\begin{split}
    &{S^j} := Y^{j} =  f^j(Y^{PA(j)^{\mathcal{G}}}, N^j)                 
    \end{split}
\end{align*}
where ${\mathcal{S}=\cup_{j \in \mathcal{E}}\{S^j\}}$ is a set of structural collections, and $PA(j)^{\mathcal{G}} \subseteq \mathcal{E}$ denotes the parents of the node ${j}$ according to the graph $\mathcal{G}$.
\end{definition}

To describe dynamic processes, again following \cite{PetJanSch17}, we extend the above definition to the dynamic case by unrolling a causal graph over time as follows:

\begin{definition}
(Dynamic SCM) \\
In analogy to a static SCM, a dynamic SCM $ \mathcal{M}({\mathcal{S}_t}, P_{N_{t}}, \mathcal{G}_t)$  is given by an acyclic graph $\mathcal{G}_t$ and exogenous noise influences $N_t^j \sim P_{N^j_{t}}$ independent over each point in time $t$ and variable $j$. $\mathcal{G}_t$ is referring to a graph consisting of an unrolled version of a summary graph $\mathcal{G}$. Following the notation of \cite{PetJanSch17}, the structural equations for node $Y_t^j$ are given by :
\begin{align*}
\begin{split}
        {S^j_t} \coloneqq Y^{j}_{t} = Y^{j}_{t-1} + f^j(Y^{PA(j)}_{t-1}, Y^{j}_{t-1}) + N^{j}_{t}
\end{split}
\end{align*}
with $PA(j)$ being the parents of node $j$ according to the summary graph $\mathcal{G}$ excluding the node itself. A notable difference from static SCMs is that the functional coupling $f$ is constant over time\footnote{Note that we restrict ourselves to additive noise in order to realize an {\em{invertible}} SCM, see \cite{pmlr-v162-budhathoki22a}}.
\label{def:dynamicSCM}
\end{definition}

\begin{definition}\label{interventional} (Interventional Dynamic SCM)
    Let $\mathcal{J}$ be a set of interventions in which each element $\xi$ can be of the following form:
    \noindent
    \begin{multicols}{2}
    \begin{equation}
        \xi := do(P_{N_t^j}) = \tilde P_{N_{t}^{j}} \label{eq:noise_intervention},
    \end{equation}\hfill\begin{equation}
         \text{ or }\qquad \xi := do(S^j_t) = \tilde S^j_t \label{eq:structural_intervention}
    \end{equation}
    \end{multicols}
    where $\tilde P_{N_{t}^{j}}$ is a new noise distribution and $\tilde S^j_t$ is a new structural equation for the node $j$ at time $t$.
    The interventional dynamic SCM is then defined by replacing either the noise distribution or structural equation within a given dynamic SCM  $\mathcal{M}({\mathcal{S}_t}, P_{N_{t}}, \mathcal{G}_t)$.
    Here  Eq. \ref{eq:noise_intervention} denotes  a soft intervention on the noise distribution whereas  Eq. \ref{eq:structural_intervention} denotes an intervention on the structural intervention. We denote the resulting intervened dynamic SCM then by   $M_{\mathcal{J}}(S_t, P_{N_t}, G_t)$.
    
\end{definition}
As each SCM (interventional or not) defines structural equations and noise distributions, it can generate a trajectory of observations. We denote the distribution of the observations generated by the SCM as $P_{\mathcal{M}}$ and the distribution of the observations generated by the intervened SCM as $P_{\mathcal{M}_{\mathcal{J}}}$. Given an observed trajectory, we can now also define the counterfactual distribution describing hypothetical trajectories which would have been observed if an (alternative) intervention had been performed.

\paragraph{Abducted  and counterfactual SCMs.} 
Let $\textbf{Y}^F$ be an observed trajectory and $\mathcal{M}$ a given dynamic SCM. In order to construct a counterfactual dynamic SCM, we define the noise posterior distribution $P_{N^{j}_{t}}(N^{j}_{t}|\textbf{Y}^F)= \delta(N^{j}_{t}-N^{F,j}_{t})$  by:
\begin{align}
N^{F,j}_{t} = - Y^{F,j}_{t-1} - f^{j}(Y^{F,PA(j)}_{t-1}, Y^{F,j}_{t-1}) + Y^{F,j}_{t} \label{eq:abduct}
\end{align}
where $f^j$ is the structural equation of the node $j$ and $PA(j)$ are the parents of the node $j$ according to the summary graph $\mathcal{G}$. The resulting dynamic SCM, in which the noise distributions $P_{N^{j}_{t}}$ are replaced with the above defined noise posterior distributions, is then denoted as $\mathcal{M}^F$ indicating that the noise distributions are abducted from the observed trajectory $\textbf{Y}^F$.
In fact, when generating trajectories from this abducted SCM, it only generates the observed trajectory $\textbf{Y}^F$ due to the above setting of the noise variables. In order to generate new counterfactual trajectories reflecting alternative outcomes, we need to perform an intervention on this abducted SCM, leading to the counterfactual SCM. That is, given an abducted SCM $\mathcal{M}^F$ and a set of interventions $\mathcal{J}$, we refer to the resulting interventional SCM $\mathcal{M}^F_{\mathcal{J}}$ as the counterfactual SCM. For example, when performing an intervention $do(P_{N_t^j})=\tilde P_{N_{t}^{j}}$ on the noise distribution at a specific point in time $t$ and a node $j$, the counterfactual SCM is defined by the following structural equations:
\begin{align}
    Y^{e}_{s} &= Y^{e}_{s-1} +  f^{e}(Y^{PA(e)}_{s-1}, Y^{e}_{s-1}) + N^{e}_{s},\qquad \text{where}\\
    N^{e}_{s} & \sim \begin{cases}\tilde P_{N_{t}^{j}} & \text{if } s = t \text{  and } e=j\\
   \delta(N^{j}_{t}-N^{F,j}_{t}) & \text{otherwise}
                      \end{cases}
\end{align}


\subsection{Root cause}
As we are interested in identifying a root cause, we state here more precisely what we mean by this term. We define a root cause as an intervention according to $M_{\mathcal{J}}(S_t, P_{N_t}, G_t)$ leading to a faulty behaviour. Here, we assume that a faulty behaviour can be detected or defined using a known classifier $\phi$. This classifier maps a time series to a binary value, indicating whether the time series is faulty. Such classifier can either be given as a known test function (e.g. corresponding to an end-of-line test in an assembly line, an assertion in a software system, or a medical diagnosis) or can be learned from data (e.g. an outlier-score function learned on normal data).

\begin{definition} (Root cause)
Given a classifier $\phi$ that determines whether an observed trajectory is faulty, we refer to a (set of) intervention(s) $\Xi$ to be the root cause of a failure associated with the classifier $\phi$, if observations  $({\bf{Y}}^F_{t, t=1,\dots T})_j$ from the  interventional SCM $\mathcal{M}_{\{ \Xi \}}$ are leading to an increased failure rate:
\begin{align*}
    \mathbb{E}_{{\bf{Y}}^F_{t, t=1,\dots T} \sim \mathcal{M}_\Xi} [\phi({\bf{Y}}^F_{t, t=1,\dots T})]-  \mathbb{E}_{{\bf{Y}}_{t, t=1,\dots T} \sim \mathcal{M}} [\phi({\bf{Y}}_{t, t=1,\dots T})] > 0
\end{align*}
\end{definition}
Note that this corresponds to the average treatment effect of an intervention on the external influence or structural intervention. If the probability of a failure for an external intervention on the noise or structure is larger than without any intervention, we assume that the failure has an underlying root cause.

\subsection{Shapley Value}
Shapley values, originally defined to quantify the contribution of individual players to the outcome of a game,  have been used by Budhathoki et al. \cite{pmlr-v162-budhathoki22a} in a static setup to define a score for nodes being potential root causes of an observed fault.  To this end, interventions (or possible root causes) are identified with players in a game whose outcome is determined by a value function that quantifyes the degree to which a set of interventions can increase the likelihood of correcting a failure (to be defined below).

\begin{definition} (Shapley value)
The  Shapley value \cite{shapley1953value} of a player $i$ out of a set $N$ of possible players to the outcome of a game characterized by the outcome function $v$ is defined by:
\begin{align*}
\begin{split}
        Sh(i) &:= \sum_{S \subseteq N \backslash \{i\}} \frac{|S|! (n - |S| -1 )!}{n!} (v(S \cup \{i\}) -v(S))
\end{split}
\end{align*}
\end{definition}
Note that in order to calculate the Shapley value, one has to sum over exponentially many subsets of the set of possible players. This is feasible only for small sets of players. As in the context of root cause analysis in a dynamic setting, the set of players corresponds to the set of possible interventions ranging over all possible times and nodes within the unrolled graph of a dynamic SCM. Due to the exponential growth of the number of possible interventions, exact Shapley value estimation is computationally infeasible for dynamic SCMs, and we have to resort to an approximate version.

\section{Method for identifying root causes}\label{sec:identification}
\begin{figure*}[ht]
    \centering
    \includegraphics[width=\textwidth]{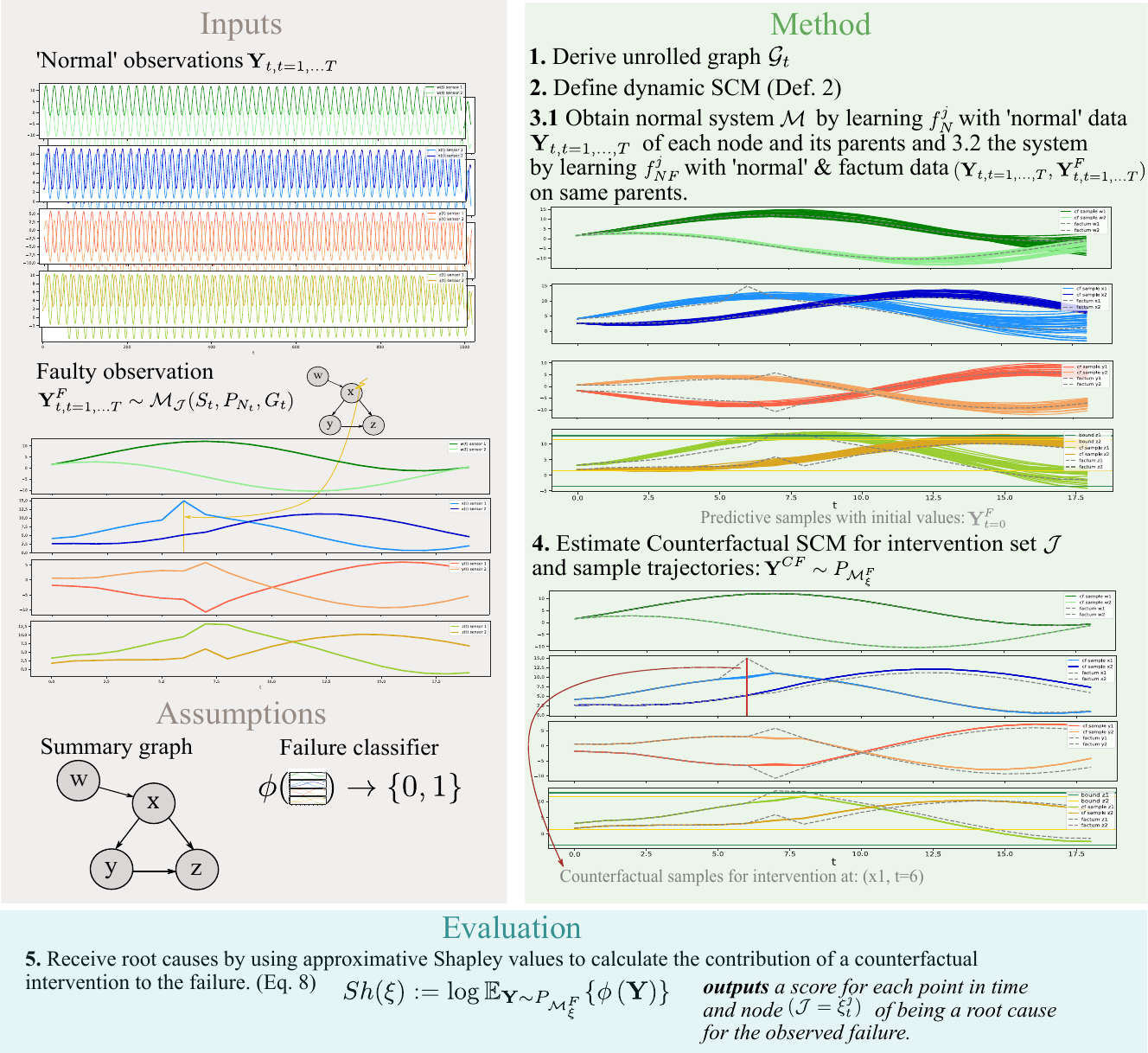}
    \caption{This figure shows an overview of the individual steps of our method.}
    \label{fig:overview}
\end{figure*} 
Now that we have the necessary background, we can describe our method for identifying root causes in dynamic SCMs. The method is based on the following steps and is illustrated in Fig.~\ref{fig:overview}.
We want to identify the root cause that caused an observed failure in a system. To this end, we cast this problem in a  counterfactual query: 
"Would the observed failure also occur if we had replaced the {\em{faulty}} behaviour of a sub-system at a certain point in time with its {\em{normal}} behaviour?".
To answer this question after we observed a faulty observation $({\bf{Y}}^F_{t, t=1,\dots T})_j$, as illustrated in \textit{Inputs} in Fig. \ref{fig:overview}, we follow the steps of counterfactual distribution calculation: abduction, action and prediction \cite{pearl2018book}. However, in order to apply those steps, we need an SCM characterizing the normal and potentially the abnormal system. To characterize the normal system, we assume to have access to data representing the normal behaviour of the system, as shown in \textit{Inputs} in Fig.\ \ref{fig:overview}. Additionally, we assume to have at least a summary graph $\mathcal{G}$ of the system. This summary graph can be obtained from expert knowledge or from data. Furthermore, as shown in \textit{Assumptions} in Fig. \ref{fig:overview}, we assume that we know a function $\phi$ that classifies an observation into faulty or normal. 

\paragraph{Fitting the model} in step 3.1 of the \textit{Method} part in Fig. \ref{fig:overview} we obtain the normal behaviour system $\mathcal{M}$ by learning the functions $f^j_{N}$ with the inputs being normal observations $Y_t$ of each node and its parents of the summary graph $\mathcal{G}$. If for both, normal as well as abnormal data, a node and hence its transition function is not anomalous, the transition function would be identical for both settings. Therefore, in 3.2 we additionally fit a transition function $f^j_{NF}$ with normal and factum data as input on the same parents and children as in 3.1 of the known graph $\mathcal{G}$ and with that we define the SCM ${\mathcal{FM}}$. We show predictive samples of $\mathcal{M}$ in the graph under 3.2.

\paragraph{Estimating the Counterfactual:} In the abduction step, we first infer the noise distribution corresponding to the observed factum. We refer to the abducted SCMs $\mathcal{M}^F$ and $\mathcal{FM}^F$ by applying the factum as function input to $f^j_{N}$ and $f^j_{NF}$ and constructing the resulting noise posterior distributions as described in Eq. \ref{eq:abduct}. We need to calculate the noise variables for both SCMs separately, because function couplings and noise variables are coupled. In the action step, we perform an intervention in $\mathcal{M}$ by $\xi_{\mathcal{M}} := \{do(P_{N_t^j}) = \tilde P_{N_t^j} \}$ (see Eq. \ref{eq:noise_intervention}), where we use the prediction error of our model to estimate the Gaussian noise variance:
    \begin{align}
            \tilde P_{N_t^j} &= \mathcal{N}(0, \sigma_{val}^2), \qquad
            &\sigma_{val}^2 = \frac{1}{V} \frac{1}{T} \sum_v \sum_t (Y^{j,v}_{t+1} - f_j(Y^{Pa(j),v}_t))^2
        \label{eq:intervention_value}
    \end{align}
with $Y^{j,v}$ being a validation trajectory of the normal data, and $V$ the number of validation trajectories. For an intervention in $\mathcal{FM}$ we intervene on the noise as before and we additionally intervene on the structure by $do(S^j_t) = \tilde S^j_t$ (see Eq. \ref{eq:structural_intervention}), which replaces the previous transition function $f^j_{FN}$ with a new structural equation $\tilde S^j_t$ consisting of the transition function $f^j_{N}$ originating from the "normal" SCM, obtained purely from training data $\xi_{\mathcal{FM}} := \{do(P_{N_t^j}) = \tilde P_{N_t^j},do(S^j_t) = \tilde S^j_t \}$.
After the construction of the corresponding counterfactual SCM we can then generate counterfactual trajectories under the different interventions $\textbf{Y}^{CF} \sim P_{\mathcal{M}^F_{\xi_{\mathcal{M}}}}$, as illustrated in 4. in Fig. \ref{fig:overview}. If an external influence on node $j$ at time $t$ leads to an abnormal factum, an intervention of the above type should remove the abnormal behaviour and therefore lead to a normal trajectory.

\paragraph{Evaluation:}
 To quantify how close  these counterfactual samples are to normal trajectories, the trajectories are processed via a classifier function $\phi$ (Eq. \ref{eq:approx-shapley}). In turn, we receive a score for each counterfactual sample indicating whether the failure was removed by the counterfactual intervention $\xi$. We then average over multiple counterfactual samples.
 To rank interventions at different times and nodes, we can use Shapley values by identifying players with interventions and match-outcomes by the average normality of the counterfactual sample.
Shapley values, however, scale exponentially and therefore we use the following simple approximation, which we obtain by ignoring interactions between different interventions, thereby only considering singleton intervention sets. Although mainly motivated by pure computational tractability, we can alternatively assume that perfectly synchronous occurrence of multiple root causes is very unlikely, thereby justifying the restriction to singleton intervention sets. Consequently we arrive the following simple expression of contribution score of individual interventions $\xi$ for each point in time and node:
\begin{align}
        Sh(\xi) &:=  \log \mathop{\mathbb{E}_{\textbf{Y}\sim P_{\mathcal{M}^F_{\xi}}}} \{ \phi\left(\textbf{Y}\right)\}  
\end{align}\label{eq:approx-shapley}
\section{Experiments} \label{sec:experiments}
In the following experiments, we evaluate the effectiveness of the proposed method for different synthetic and real world data-sets. As for synthetic data-sets, we consider both linear and non-linear dynamic systems with single point external failure-causing disturbances as well as a benchmark data-set for identifying structural causes for anomalies \cite{pmlr-v206-assaad23a}. As for the real-world data-set, we are investigating dynamic water flow rate in rivers \cite{RiverDataSource}. For our synthetic experiments, we perform two meta-experiments which analyze the influence on the model performance of varying root cause injections and how robust the model is against violating the assumption that the causal graph is known.
We denote our models, a linear and a non-linear model both performing a counterfactual intervention on the external noise influence and on the structural equation with \textit{Lin($S^j_t,N_t^j$)} and \textit{NLin($S^j_t,N_t^j$)}. We compare against a linear layer model with counterfactual noise-influence intervention \textit{Lin($N_t^j$)}, similar to \cite{pmlr-v162-budhathoki22a} as well as against EasyRCA \cite{pmlr-v206-assaad23a} in the benchmarking experiment. For completeness, we additionally provide a nonlinear model \textit{NLin($N_t^j$)} with counterfactual noise-influence intervention. In order to model the non-linear dynamic SCM, for \textit{NLin} we use a simple three-layer residual neural network (ResNet) with hyperbolic tangent activation functions and 128 neurons as latent layer. 

\subsection{Experimental datasets}

\paragraph{Linear synthetic system:}
In our first data-set, we consider a linear multivariate system with additive Gaussian noise consisting of four nodes ($w,x,y,z$), each having two dimensions. The summary graph of the system is shown in \textit{Assumptions} in Fig. \ref{fig:overview}. The structural equations of the system are of the form:
\begin{align*}
    Y^{j}_{t} := A^i \textbf{Y}^j_{t-1} + \sum_{k \in PA(j)}B^{k} \textbf{Y}^{k}_{t-1} + C^{l}N_t^j, \qquad (N_t^j)_d \sim \mathcal{N}(0,1) \quad \forall d 
\end{align*}
with $N_t^j$ being zero mean standard Gaussian noise. For this system we chose the transition matrices such that they generate a stable system by using eigenvalues smaller than 1 (see Appendix). 
To simulate a root cause, we inject an additive constant term at a single dimension of a node $j$ at time $t$ to the equation above. Instead of a learned anomaly scoring function, in this experiment, we assume to have access to a function that checks the validity of a given observation, similarly as it would be in a manufacturing scenario, in which an end of line test is performed \cite{göbler2024causalassembly}. Therefore, we examine if a failure on the "last" node in a manufacturing line (here "last" node in the summary graph is $z$) has occurred. To this end, we use a threshold function, fixed over time for each dimension of node $z$. More precisely, this classifier can be applied to any time-series observation $({\bf{Y}}_t^j)_{t\in \{1,...,T\}, \,j\in \{w, x,y,z\}}$: 
\begin{align*}
        \phi({\bf{Y}}) = 1- \frac{1}{D_z} \sum_{k=1}^{D_z}\mathds{1}_{[(\mu_z)_k -(\sigma_z)_k, (\mu_z)_k+(\sigma_z)_k]}(\textbf{Y}^z_k)
\end{align*}
Here, the dimension of node $z$ is denoted with $D_z$. Note that this function provides a gradual feedback of how many of the dimensions in node $z$ are outside of the pre-specified corridor given by the threshold function.

\paragraph{FitzHugh–Nagumo system:}
Next, to allow for non-linear dynamic behaviour, we are generating data of the FitzHugh–Nagumo system (FHN), which is cyclic with regard to its summary graph, but acyclic in the unrolled graph $\mathcal{G}_t$. Although being a multivariate system, as the two dimensions interact, the corresponding dynamic SCM consists of one node $x$ with two dimensions:
\begin{align*}
    \dot{x_1} &= 3(x_1 - x^3_{1}/3 + x_2),\qquad    &\dot{x_2} = (0.2 - 3x_1 - 0.2x_2)/3
\end{align*}
We chose the initial values as in \cite{hegde2022variational} but with slightly reduced additive Gaussian noise variance $\sigma^2 = 0.0025$. The root cause is simulated similarly to the linear system by adding a constant to the difference equation at one dimension and time point. We classify an observation as faulty, if it deviates too much from a normal observation. As we have, in this setting, access to the ground truth, a normal observation is represented by a trajectory generated from the ground truth system. Consequently, the classifier consists of a time-varying threshold bound around each dimension of the normal observation without the injected root cause of node $x$. Denoting the expected trajectory from the system by ${\bf{E}}$ a given observation ${\bf{Y}}$ is then classified to be faulty if it does not deviate more than 10 standard deviations at any point in time from the expected trajectory: $
        \phi({\bf{Y}}) = 1- \prod_t\mathds{1}_{[{\bf E}^x_t-10\sigma_x, {\bf E}^x_t +10 \sigma_x]}({\bf{Y}}^x_t)
     $.

\subsection{Evaluation}
\begin{figure}
    \centering
    {\includegraphics[width=0.49\textwidth]{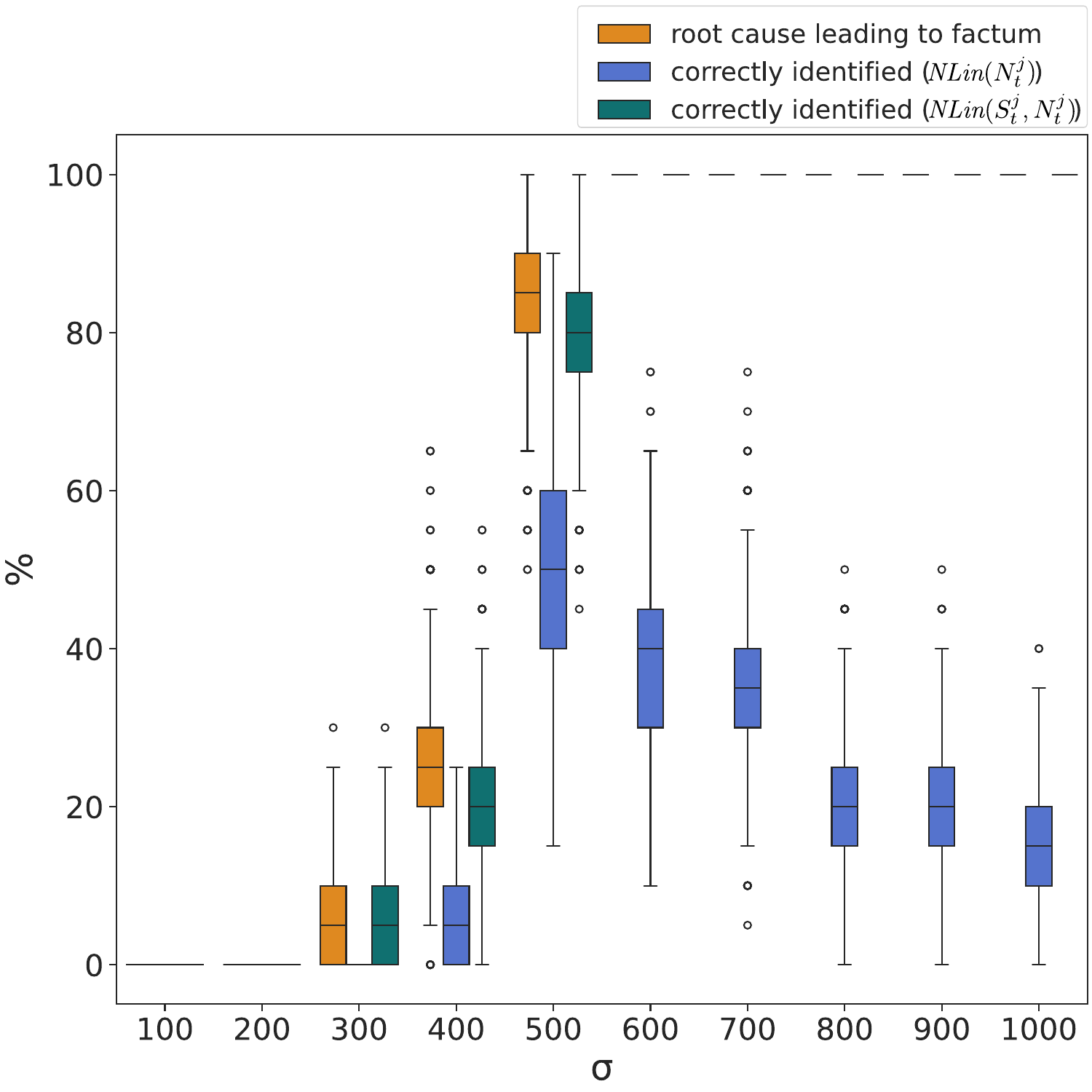}\label{fig:nl_model_lin_system_boxplot}}
    {\includegraphics[width=0.49\textwidth]{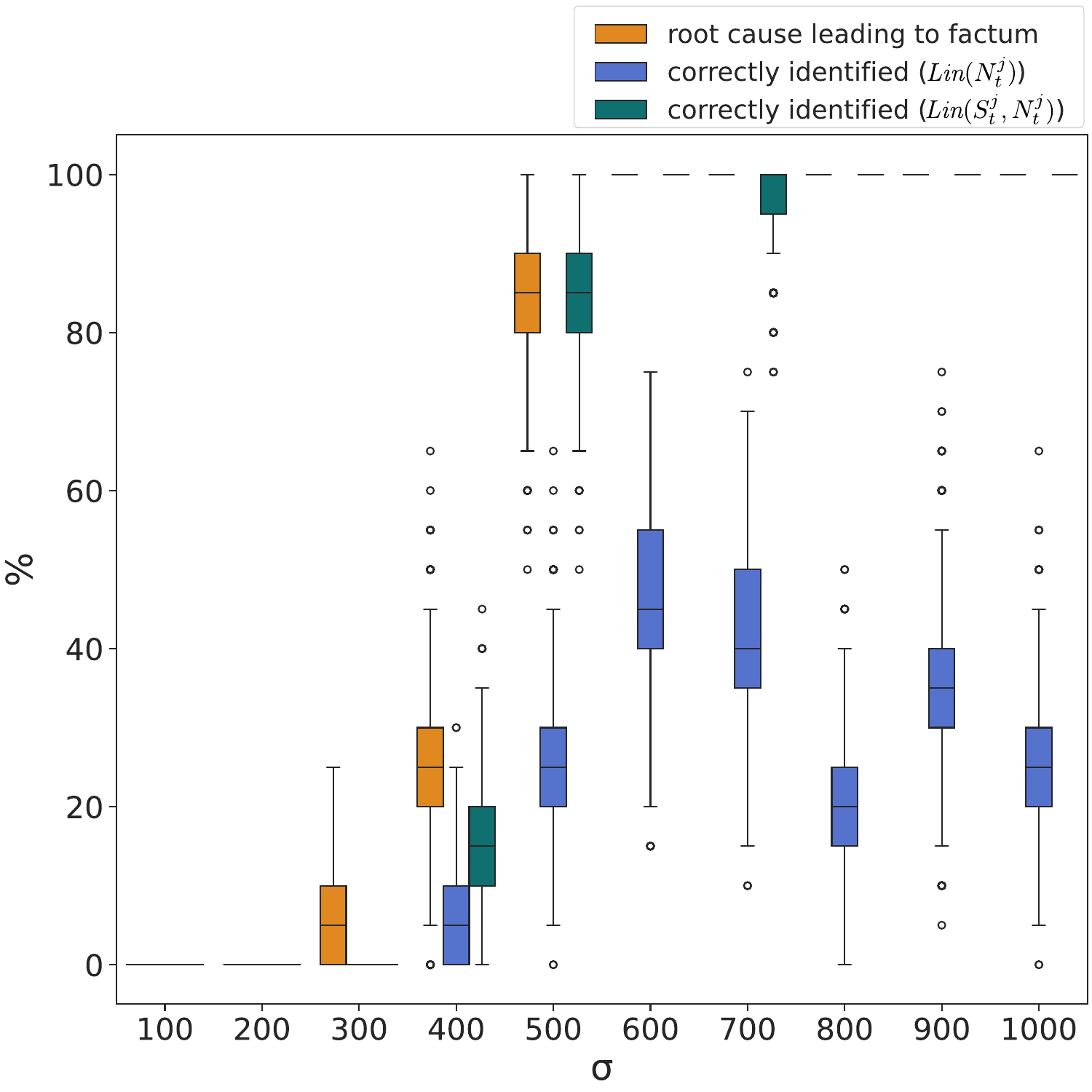}\label{fig:lin_model_lin_system_boxplot}}
\caption{The root cause was injected at a random node $j=x_1$ at $t=6$ with varying constants in $[1,10]$. The horizontal axis shows the injected constant in relation to the noise standard deviation denoted by $\sigma$. We report how many root causes could be identified in \%.}
\label{fig:injection-boxplot}
\end{figure}

\begin{figure}
    \centering
    \includegraphics[width=\linewidth]{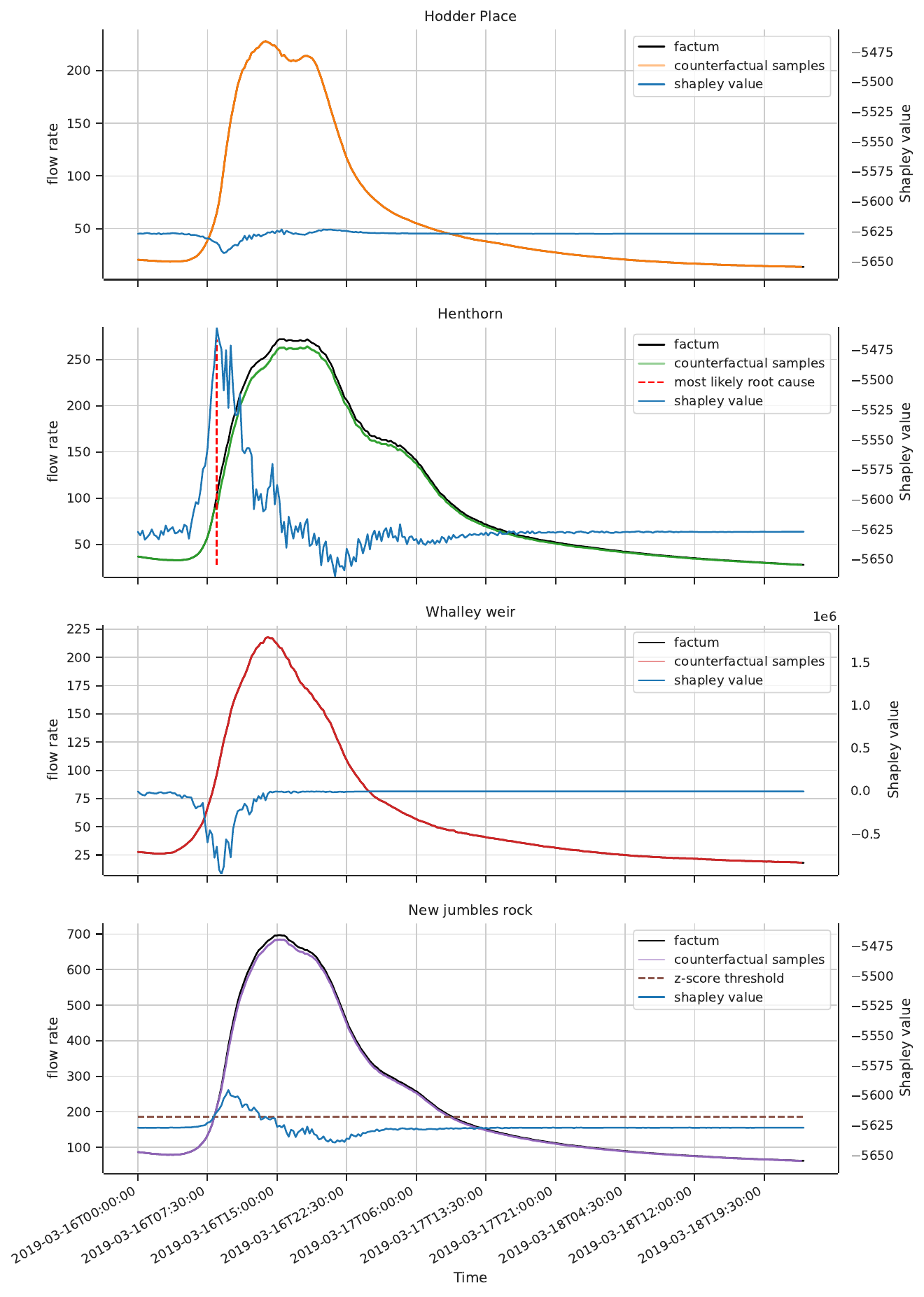}
    \caption{We show five counterfactual samples (for each station) of our model \textit{NLin($S^j_t,N_t^j$)} with the intervention at the predicted root cause at 08:30 on 16.03.2019. Additionally, we illustrate the resulting Shapley values for each time point, showing that right before the failure occurs the Shapley values increase.}
    \label{fig:river-result}
\end{figure}

When we have drawn counterfactual samples from our model, we calculate the approximate Shapley values (see Eq. \ref{eq:approx-shapley}) and use the $\phi$ function to evaluate each performed intervention based on whether it corrected the failure. The root cause is the intervention of the node $j$ at time $t$ that has the highest influence on the failure. 
If all counterfactual samples lead to the same $\phi$ evaluation for all interventions, then no unique root cause could be identified. However, due to random sampling of the counterfactual, this is an unlikely scenario (see for example Fig.\ref{fig:river-result}.) Nevertheless, for the evaluation, we only require that the ground truth root cause is within the set of identified root causes.
\begin{figure}
    \centering
    \includegraphics[width=0.95\linewidth]{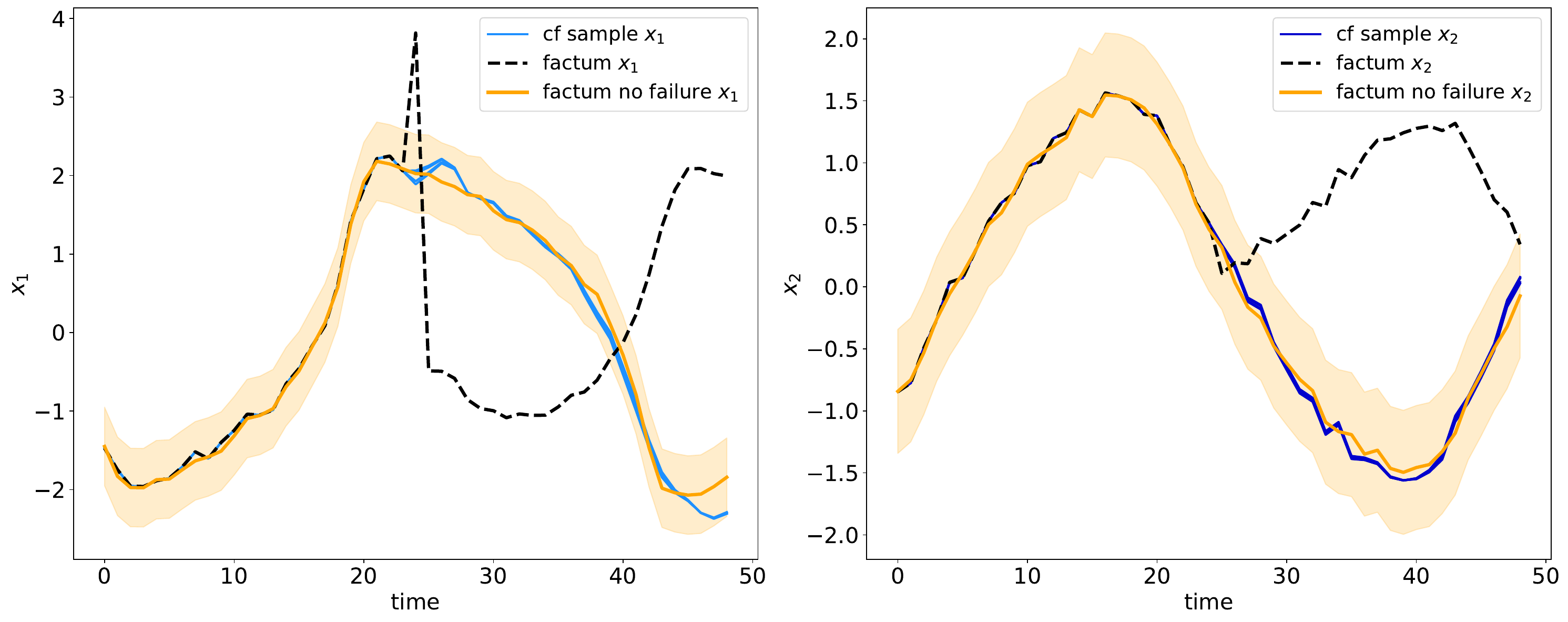}
    \caption{The figure shows the counterfactual samples for the FHN system with injected root cause at ($j=x_1,t=24$). The injected root cause disrupts the system observation heavily (black dashed line). However, the counterfactual intervention performed by our model \textit{NLin($S^j_t,N_t^j$)} corrects the failure in both dimensions, such that it lies inside the threshold region (orange area).}
    \label{fig:cf_FHN}
\end{figure} 
In Fig. \ref{fig:cf_FHN} we show five counterfactual samples for the nonlinear FHN system at the actual root cause injection point. Although the injected root cause is fairly large with regard to the interval of the normal observation without failure (drawn as orange line), the counterfactual intervention performed by our model \textit{NLin($S^j_t,N_t^j$)} corrects the failure for both dimensions of $x$. In order to analyze root cause injections and how the identification capabilities of our model behave under varying injections, we performed an \textit{Injection experiment} for the synthetic linear and nonlinear FHN system. External disturbances in dynamic systems may be propagated and thereby increase their impact. Alternatively, if the system is robust against incremental noise (as it is the case in the defined systems above due to the external noise influence even under the 'normal' conditions), it is not obvious how large an external influence at which point in time is noticeable. In Fig. \ref{fig:injection-boxplot}, we show varying root cause injections for the linear synthetic system (varying constant added to the structural equation) over 20 randomly sampled facta with $T=20$. It can be seen that the models intervening on the structure and the noise achieve a significantly higher identification score for large added constants. This could be due to a large root cause, in this setting leading to a factum with high distance to the normal data, which may lead to a divergence over time of the normal behaviour system $\mathcal{M}$. Note that we did a similar injection experiment for the FHN system, which can be found in the Appendix.
\paragraph{Assumption violation.}
We probe our models on violation of the causal graph assumption for the linear synthetic system. For this, we modify the causal graph used by the underlying model through adding or removing random edges, while keeping the original summary graph for data generation. We use the same facta generated as $\sigma$ = 500 in Fig. \ref{fig:injection-boxplot}. In Tab. \ref{tb:results_robustness} it can be seen that removing edges for all models has a stronger impact on predictive performance than adding. As expected, \textit{Lin($(S^j_t,N_t^j)$)} performs best on this linear system, closely followed by \textit{NLin($(S^j_t,N_t^j)$)}. It must be mentioned that in a graph with only four edges, removing an edge is a major incision in the model assumption. 

\begin{table*}[ht!]
\centering
 \caption{We show the \textbf{Accuracy} for the setup $\sigma$ = 500 of the linear synthetic system (see Fig. \ref{fig:injection-boxplot}) with varying number of \textit{removed or added edges} of the summary graph $\mathcal{G}$ used by the models.}
\begin{tabular}{lcccccl}\toprule
           & $NLin(S^j_t,N_t^j)$    & $NLin(N_t^j)$  
           & $Lin(S^j_t,N_t^j)$     & $Lin(N_t^j)$ \\\midrule
nr. of removed edges &&&& \\
$\text{1}$   & 0.47 $\pm$ 0.25   & 0.23 $\pm$ 0.18  & 0.47 $\pm$ 0.24             & 0.18 $\pm$ 0.15   \\
$\text{2}$   & 0.29 $\pm$ 0.20   & 0.06 $\pm$ 0.06  & 0.47 $\pm$ 0.24            & 0.12 $\pm$ 0.10      \\\midrule
nr. of added edges &&&& \\
$\text{1}$      & 0.82 $\pm$ 0.15   & 0.12 $\pm$ 0.10  & 1.0 $\pm$ 0.0           & 0.18 $\pm$ 0.15         \\
$\text{2}$      & 0.88 $\pm$ 0.10   & 0.0 $\pm$ 0.0    & 0.88 $\pm$ 0.10           & 0.06 $\pm$ 0.06           \\\midrule
\end{tabular}
\label{tb:results_robustness}
\end{table*}

\paragraph{Linear EasyRCA Benchmark.}
We compare against the linear univariate benchmark of \cite{pmlr-v206-assaad23a} consisting of six nodes and two types of root causes. The parametric root cause meaning they change the coefficient of the parent nodes to a random uniform sampled value. As a special case of the parametric setting, they inject structural root causes, which set the coefficient of the parent nodes to zero. Since EasyRCA excludes single time point root causes, in order to do a fair comparison, we only rank sets of interventions, where we intervene on all times for a given node and evaluating it accordingly by $Sh({\xi_0^j,...\xi_T^j})$. 
In their work they inject on two nodes, where one is always the root node of the system and the other one a randomly chosen node. As in their benchmark comparison the root node root cause is excluded, we exclude it from the evaluation as well. 
In the evaluation they distinguish for parametric and structural root causes, but because our model makes no prediction about the type of root cause, it is sufficient if EasyRCA predicted root causes contain the true root cause, regardless of the type. To rate the normality of a given trajectory ${\bf{Y}}$, we make use of the learned dynamical SCM $\mathcal{M}$ which was fitted on normal observations of the system. More precisely, for the EasyRCA benchmark as well as the following River experiment, we used an outlier score similarly to \cite{pmlr-v130-budhathoki21a}, based on the learned dynamic SCM. That is, given a dynamic SCM $\mathcal{M}$ consisting of $N$ nodes and providing the conditional distribution $p({\bf{Y}}_t^j|{\bf{Y}}_t^{PA(j,t)})$ via the dynamics equation learned from normal observational data $({\bf{Y}})_k$, we can define the following outlier score:
\begin{align}
    \phi({\bf{Y}}) = \frac{1}{NT} \sum_{j,t} \log p({\bf{Y}}_t^j|{\bf{Y}}_t^{PA(j,t)}) \label{eq:phi_easyrca_river}
\end{align}
In Tab.\ \ref{tb:results_dynamic}, it can be seen that in general the intervention $(S^j_t,N_t^j)$ is preferable to an intervention only on $(N_t^j)$. For the linear systems, the accuracy of \textit{NLin$(S^j_t,N_t^j)$} and \textit{Lin$(S^j_t,N_t^j)$} are similarly good, while \textit{EasyRCA} shows lower performance in the factum $T=100$ experiments. However, \textit{Lin$(S^j_t,N_t^j)$} is inadequate for addressing the complexities of the nonlinear problem.


\begin{table*}[ht!]
\centering
 \caption{We report the {\bf{Accuracy}} over 20 facta of the summary graph on a linear system and the FHN oscillator. In the lower part of the table we present the experimental results of the EasyRCA benchmark \cite{pmlr-v206-assaad23a} comparing the accuracy for one factum over 30 graphs for different factum lengths $T$ (here, normal data has the same size $T$). \protect\footnotemark}
 \resizebox{1\textwidth}{!}{
\begin{tabular}{lcccccl}\toprule
& \multicolumn{2}{c}{NLin} & \multicolumn{2}{c}{Lin} & \multicolumn{2}{c}{EasyRCA}\\
            \cmidrule(lr){2-3}\cmidrule(lr){4-5}\cmidrule(lr){6-7}
           & $(S^j_t,N_t^j)$    & $(N_t^j)$ 
           & $(S^j_t,N_t^j)$    & $(N_t^j)$  & \\\midrule
$\text{Lin. system}$    & 0.94 $\pm$ 0.05   & 0.59 $\pm$ 0.24  & 1.0 $\pm$ 0.0 & 0.29 $\pm$ 0.20 & -  \\
$\text{FHN oscillator}$   & 0.90 $\pm$ 0.09                & 0.65 $\pm$ 0.23                & 0.20 $\pm$ 0.16           & 0.15 $\pm$ 0.13             &- \\\midrule
$\text{Lin. Parametric}$ \\
\quad $\texttt{Factum-100}$  & 1.0 $\pm$ 0.0   & 0.97 $\pm$ 0.03   & 1.0 $\pm$ 0.0     & 0.97 $\pm$ 0.03  & 0.87 $\pm$ 0.12\\
\quad $\texttt{Factum-200}$  & 0.97 $\pm$ 0.03 & 0.93 $\pm$ 0.06   & 1.0 $\pm$ 0.0     & 0.93 $\pm$ 0.06  & 0.93 $\pm$ 0.06\\
\quad $\texttt{Factum-500}$  & 1.0 $\pm$ 0.0   & 0.97 $\pm$ 0.03   & 1.0 $\pm$ 0.0     & 0.97 $\pm$ 0.03  & 1.0 $\pm$ 0.0  \\
\quad $\texttt{Factum-1000}$ & 1.0 $\pm$ 0.0   & 1.0 $\pm$ 0.0     & 0.97 $\pm$ 0.03   & 0.83 $\pm$ 0.14  & 0.97 $\pm$ 0.03  \\
$\text{Lin. Structural}$ \\
\quad $\texttt{Factum-100}$  & 1.0 $\pm$ 0.0    & 0.87 $\pm$ 0.12   & 1.0 $\pm$ 0.0    & 0.83 $\pm$ 0.14    & 0.8 $\pm$ 0.16   \\
\quad $\texttt{Factum-200}$  & 0.90 $\pm$ 0.09  & 0.27 $\pm$ 0.20   & 0.70 $\pm$ 0.21  & 0.53 $\pm$ 0.25    & 0.90 $\pm$ 0.09  \\
\quad $\texttt{Factum-500}$  & 1.0 $\pm$ 0.0    & 1.0 $\pm$ 0.0     & 0.87 $\pm$ 0.12  & 1.0 $\pm$ 0.0      & 1.0 $\pm$ 0.0  \\
\quad $\texttt{Factum-1000}$ & 1.0 $\pm$ 0.0    & 0.8 $\pm$ 0.16    & 1.0 $\pm$ 0.0    & 1.0 $\pm$ 0.0      & 0.97 $\pm$ 0.03  \\
\bottomrule
\end{tabular}
\label{tb:results_dynamic}}
\end{table*}
\footnotetext{We excluded the 2000 factum length experiment of the EasyRCA benchmark for computational reasons. Additionally, note that since EasyRCA is univariate, it can not be applied to our synthetic systems.}

\paragraph{Real World River Experiment.}

We analyse our method on real-world data 
\begin{figure}
    \centering
    \includegraphics[width=0.45\linewidth]{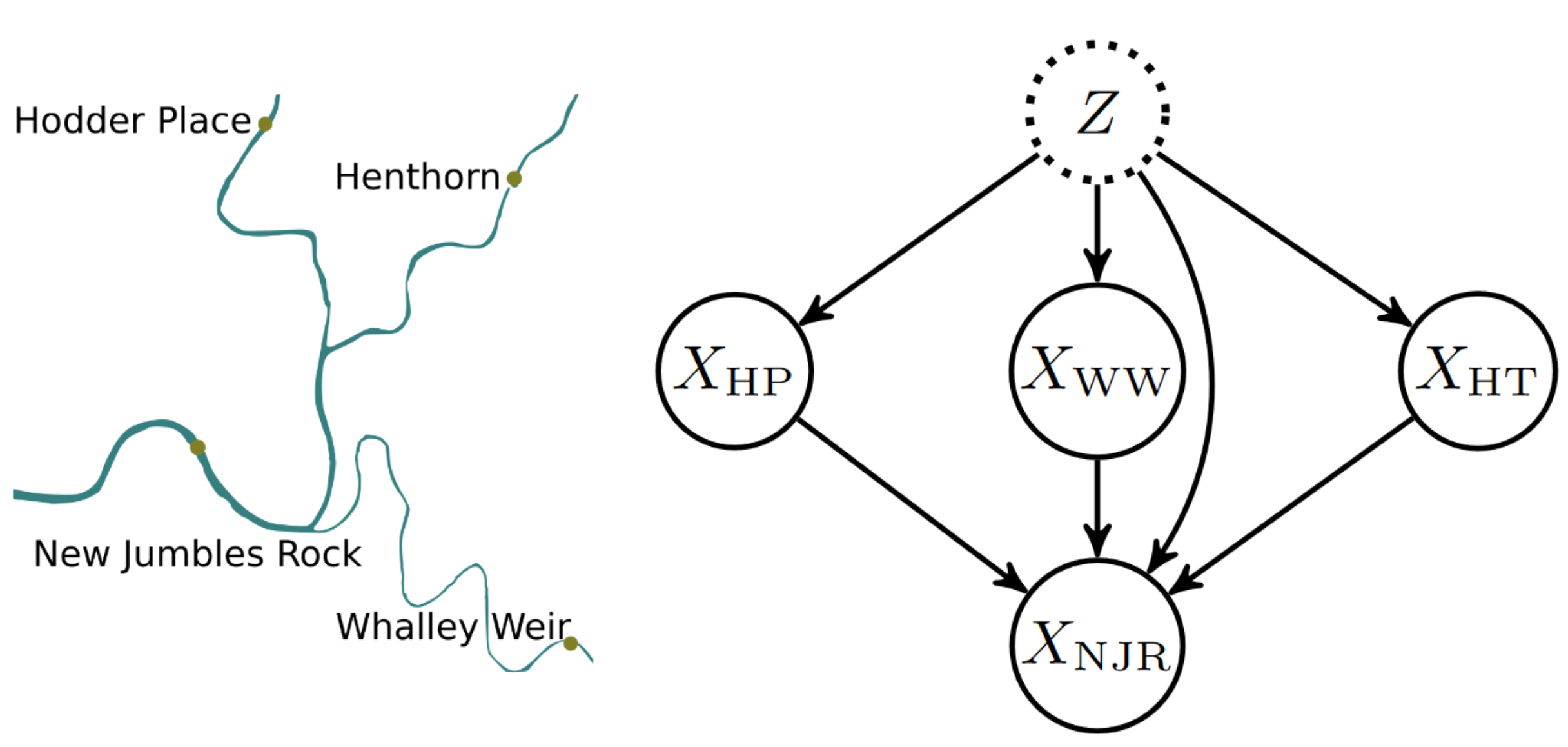}
    \caption{With the geographical knowledge of the river flow, a summary graph can be inferred (Figure taken from \cite{pmlr-v130-budhathoki21a}).}
    \label{fig:river-summary-graph}
\end{figure} 
considering a univariate river experiment consisting of four nodes. The nodes represent measuring stations of the Ribble River in England (data is from \cite{RiverDataSource}).
These measuring stations are influenced by unknown external influences such as for example rain. For this reason, the summary graph includes an unobserved confounder $Z$ that influences all nodes. This unobserved confounder affects the accuracy of our model when learning the normal system $\mathcal{M}$ from observational data.
The nodes represent stations of the Ribble River that measure the flow rate. Although this data-set has been investigated in \cite{pmlr-v130-budhathoki21a}, as a result of our dynamic viewpoint, we consider a slightly different factum. They consider four time points as static facta and infer the root causes for these. In contrast, we consider an entire time series as factum and infer the root cause. In addition, we use a finer time resolution of 15-minute intervals instead of averaged daily values, which has the advantage that the resulting SCM is less prone to instantaneous effects due to aggregation within a time-window. The finer resolution means that we consider a shorter period of time, namely the three days from 16.03.2019 to 19.03.2019 in which the flow rate is particularly high. As training data, we use the same time span as \cite{pmlr-v130-budhathoki21a} from 01.01.2010 to 31.12.2018. They provide a z-score threshold for the New jumbles rock station, which we use as $\phi$ in \ref{eq:phi_easyrca_river}, see also Fig.\ref{fig:river-result}. 
We find the Shapley values with the highest scores at station Henthorn, which is an upstream station of the New jumbles rock station. Although no ground truth root cause exist for this experiment since it is a real world example, the result is plausible both geographically and with regard to the time point. Nevertheless, the counterfactual intervention cannot correct the failure, as  the counterfactual sample is not below the z-score threshold. This could be due to the fact that the influence of the unobserved confounders is particularly high.

\section{Conclusion}\label{sec:conclusion}
In this paper, we have presented a method for identifying root causes in dynamic systems based on counterfactual reasoning. As the proposed method ranks individual interventions corresponding to individual nodes or sensors at particular times within a trajectory, our method is capable of exploiting not only the causal structure but also the natural direction of causality over time. By modelling temporal transitions with a non-linear neural network and a Shapley value approximation, we are able to remove important limitations of current counterfactual root cause analysis methods.
While we demonstrated both on synthetic as well as real data the effectiveness of our method in identifying root causes in dynamic systems, there are several directions for further improvement. For example, our method is current limited to the assumption that the root cause consists of a single intervention and that the causal graphical structure is known as well as the absence of latent confounders. In future work, we plan to extend our method to identify multiple root causes and to include uncertainties in the graphical structure as well as potential latent confounders.

\begin{credits}

\subsubsection{\discintname}
The authors have no competing interests to declare that are
relevant to the content of this article.
\end{credits}
%
%
%
\bibliographystyle{splncs04}
\bibliography{bib/main}

\end{document}